\documentclass[letterpaper, 10 pt, conference]{ieeeconf}  

\IEEEoverridecommandlockouts                              

\usepackage{CJKutf8}
\usepackage{amsmath} 
\usepackage{amssymb}  
\usepackage{graphicx}
\usepackage{color}
\usepackage{xurl}
\usepackage[hidelinks]{hyperref}
\usepackage{cuted}
\usepackage{capt-of}
\usepackage{tabularx}
\usepackage{booktabs}   
\usepackage{caption}
\usepackage{siunitx}
\usepackage{multirow}
\usepackage{bm}
\usepackage{makecell}

\usepackage{graphicx}

\DeclareRobustCommand{\cheetahemoji}{\raisebox{-0.15em}{\includegraphics[height=1.1em]{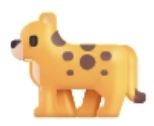}}}

\usepackage{diagbox}

\title{\LARGE \bf
S-Cheetah: A Novel Quadrupedal Robot with a 3-DOF Active Spine Learning Agile Locomotion
}

\author{Zimu Li and Weibang Bai$^*$
\thanks{Zimu Li and Weibang Bai are with the School of Information Science and Technology, ShanghaiTech University, Shanghai 201210, China}
\thanks{*Corresponding author: {\tt\small wbbai@shanghaitech.edu.cn}}
}
\begin{document}

\maketitle
\thispagestyle{empty}
\pagestyle{empty}


\begin{strip}
    \vspace*{-2cm}
    \centering
    \includegraphics[width=1.0\textwidth]{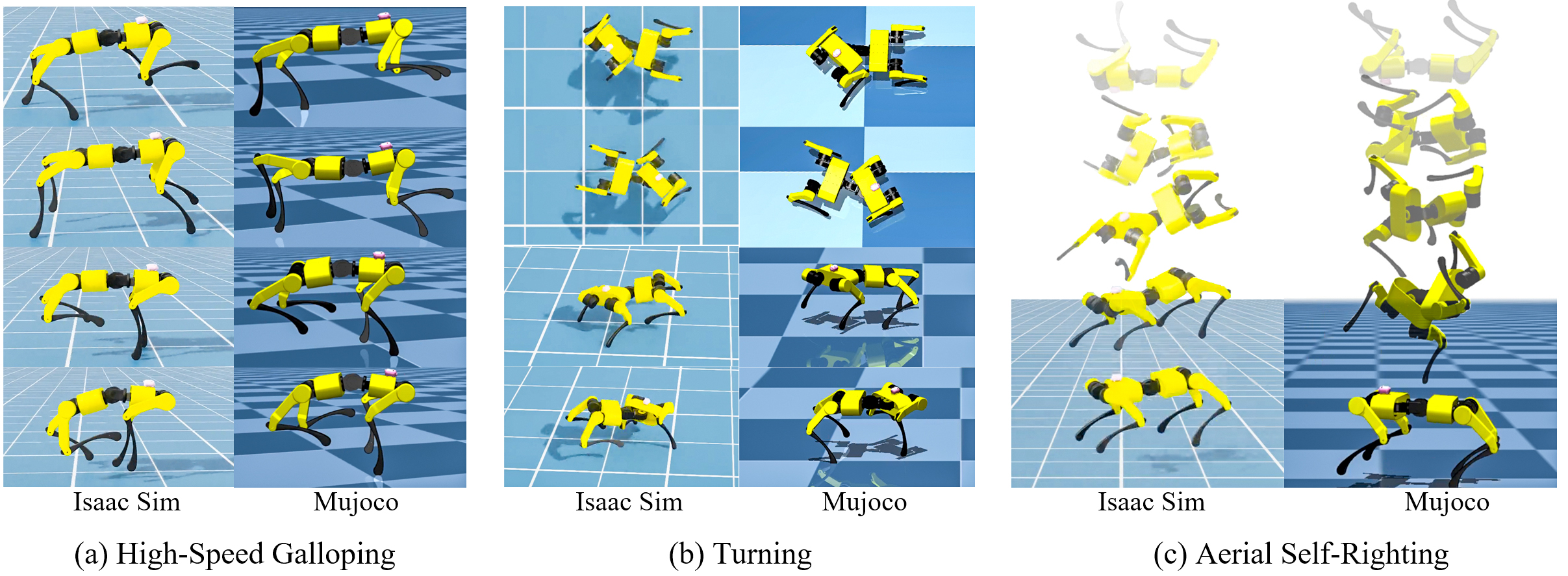} 
    \vspace{-0.3cm}
    \captionof{figure}[S-Cheetah Teaser]{\textit{S-Cheetah} \protect\cheetahemoji \textit{ : A 3-DOF \underline{S}pine-driven \underline{Cheetah}-inspired Quadrupedal Robot for Agile Locomotion}. Our proposed S-Cheetah demonstrates agile locomotion capabilities across diverse scenarios. (a) High-speed rotary G2 galloping. (b) Maneuverable in-place turning and moving turning. (c) Emergent feline-inspired aerial self-righting.}
    \label{voltah}
    \vspace{-0.3cm}
\end{strip}

\begin{abstract}
The biological spine of quadrupeds enables sagittal flexion/extension, lateral bending, and axial rotation, playing a crucial role in highly agile and dexterous locomotion. While numerous studies have integrated active spinal joints into quadrupedal robots to enhance agility, most designs simplify control complexity by reducing spinal degrees of freedom (DOF), failing to achieve the spatial tri-axial rotation characteristic of biological spines. Consequently, replicating a multi-DOF biomimetic spine and effectively leveraging it to empower the agile locomotion of quadrupedal robots remains a significant research challenge. In this study, we present S-Cheetah, a quadrupedal robot featuring a 3-DOF bio-inspired serial active spine capable of biomimetic spatial tri-axial rotation. To empower the robot to fully utilize this active spine, we developed a specialized reinforcement learning framework to
actively promote the engagement of the introduced spine and maximize the robot's locomotive capabilities by integrating an acceleration curriculum learning strategy with tailored reward functions, such as a gallop gait reward, a spine undulation reward, and a spine steering reward. Experimental results demonstrate that S-Cheetah can achieve a peak speed of $6.9\ \text{m/s}$ using the rotary G2 gallop gait and an in-place turning rate of $7.2\ \text{rad/s}$. Besides, the system exhibits an emergent, feline-inspired aerial self-righting capability, allowing it to land stably on four feet from arbitrary orientations during free fall. Finally, through extensive evaluations across diverse locomotion tasks, we prove that the introduction of the proposed 3-DOF spine comprehensively enhances the locomotive agility of quadrupedal robots. Project website: himmy-robotics.github.io/scheetah
\end{abstract}

\section{Introduction}
Through hundreds of millions of years of evolution, terrestrial quadrupeds have developed a sophisticated synergy between their limbs and spinal column, endowing them with extraordinary terrain adaptability and locomotor agility \cite{alexander2003principles}. This is manifested in their diverse gaits, from the walk and trot to the high-speed gallop \cite{hildebrand1989quadrupedal}. Throughout agile locomotion, the spine serves not merely as a structural support but as a central engine for power modulation. Particularly during galloping, the rhythmic flexion and extension of the spinal joints significantly amplify stride length, providing substantial gains in dynamic performance \cite{kamimura2021dynamical}.

Inspired by these biological archetypes, the field of quadrupedal robotics has made remarkable progress. The integration of high-torque-density actuators \cite{mini-cheetah} and reinforcement learning-based control \cite{doi:10.1126/scirobotics.aau5872} has enabled modern quadrupedal robots to achieve unprecedented agility, including complex parkour maneuvers in challenging environments \cite{doi:10.1126/scirobotics.adi7566, doi:10.1126/scirobotics.ads6192}. Notably, the ``Black Panther'' quadrupedal robot achieved a top speed of $10.3\ \text{m/s}$ \cite{Xinhua2025BlackPanther}, and the ``White Rhino'' set a Guinness World Record for the fastest $100\ \text{m}$ by a quadrupedal robot at $16.33$ seconds \cite{GWR_WhiteRhino}. However, in the pursuit of control simplicity and structural robustness, these high-performance platforms universally adopt a rigid-trunk design, omitting the degrees of freedom (DOF) provided by a biological spine. Consequently, even at peak velocities, they predominantly rely on the highly stable yet dynamically conservative trot gait rather than the gallop, which is known for its explosive power and speed in nature \cite{hildebrand1989quadrupedal}, fundamentally limiting further improvements in robotic dynamic performance.

To bridge this gap, numerous studies have explored integrating articulated spinal joints into quadrupedal robots. Bhattacharya et al. \cite{8956332} designed a quadrupedal robot with a 2-DOF pitch-actuated spine, achieving a running speed of $2.21\ \text{m/s}$ in simulation via reinforcement learning. Caporale et al. \cite{caporale2023twisting} developed a quadrupedal robot featuring a single roll-DOF spine, experimentally demonstrating that spinal twisting during locomotion reduces energy consumption. Similarly, Qian et al. \cite{qian2024robust} designed a quadruped with a roll-DOF spine, showing through experiments that the introduction of spinal rolling reduces the turning radius while enhancing path-tracking capability and adaptability to complex terrain. Furthermore, Yoneda et al. \cite{11245909} utilized reinforcement learning to enable a quadrupedal robot with a single pitch-DOF spine to perform wall-climbing maneuvers. The ``Yat-sen Lion II'' robot presented by Yang et al. \cite{11246577} features a 2-DOF spine (pitch and yaw), demonstrating that spinal integration results in a smaller turning radius and more agile trajectory tracking performance. Some research has explored higher DOFs. Wang et al. \cite{10610113} employed a symmetric five-bar mechanism to construct a 4-DOF spine capable of twisting, extension, contraction, and rotation. However, it lacks the extensive range of motion found in biological quadrupeds, and the spine functions as a passive spring rather than an actively controlled component. Wu et al. \cite{WuY2-RSS-25} introduced a novel spinal structure where each module possesses two DOFs, allowing for serially connected modules to increase the total DOF, and demonstrated aerial self-righting during free-fall through spinal articulation. However, since each module is constrained to planar pitch and yaw motions, the assembled spine still lacks a roll DOF. Moreover, the complex multi-module spinal design coupled with the simplified leg structure limits the robot's ability to perform other complex locomotion tasks. Regarding speed performance, the robot developed by Hu et al. \cite{10907590} achieved a maximum galloping speed of $4.9\ \text{m/s}$ in simulation, yet the pitch spinal joint exhibited only negligible angular variation. The ``Yat-sen Lion'' robot introduced by Li et al. \cite{10160896} attained a top bounding speed of $4.0\ \text{m/s}$ in simulation. Its two pitch spinal joints (front and rear) reached a maximum angle of $0.21\ \text{rad}$ ($\approx12.0^\circ$)---equivalent to a single joint angle of $0.42\ \text{rad}$ ($\approx24.1^\circ$). However, the spinal angles reached during flexion and extension were identical, which is inconsistent with the asymmetric spinal motion patterns observed in biological quadrupeds during running. For instance, a galloping cheetah's spine achieves a maximum flexion angle of $84^\circ$ versus an extension angle of only $44^\circ$ \cite{belyaev2024running}.

The unique biological structure of the quadrupedal spine enables spatial tri-axial rotation, which plays a crucial role not only in enhancing locomotive agility but also in enabling remarkable behaviors such as feline aerial self-righting during falls \cite{kane1969dynamical}. However, current research on spinal quadrupedal robots rarely achieves this bio-inspired mode of tri-axial rotation, and although Wu et al. \cite{WuY2-RSS-25} demonstrated spinal self-righting, no existing work has realized both agile locomotion and aerial self-righting within a unified framework. Furthermore, regarding control methods, whether using reinforcement learning \cite{8956332, 11245909, 10610113} or Model Predictive Control (MPC) \cite{11246577, WuY2-RSS-25, 10907590, 10160896}, a major challenge remains: how to effectively utilize the spine to holistically enhance the locomotive agility of quadrupedal robots.

To address these challenges, we propose S-Cheetah: a novel quadrupedal robot featuring a bio-inspired serial 3-DOF active spine. This spine is capable of spatial tri-axial rotation, enabling it to accurately replicate the spinal kinematics of biological quadrupeds. Furthermore, we developed a specialized reinforcement learning framework with tailored reward functions that effectively synergizes spinal articulation with limb movements, thereby achieving superior locomotive agility.

The main contributions of our work are summarized as follows:
\begin{enumerate}
  \item We develop a novel quadrupedal robot integrated with a $3$-DOF bio-inspired serial active spine, capable of accurately replicating the complex tri-axial spinal rotations observed in biological quadrupedal locomotion.
  \item We propose a reinforcement learning framework with tailored reward functions that effectively coordinates spinal and limb motions, enabling the robot to achieve high-speed galloping, maneuverable turning, precise path-tracking, and emergent feline-inspired aerial self-righting.
  \item We conduct comprehensive evaluations across multiple highly agile locomotion tasks in both Isaac Sim and MuJoCo simulation environments, demonstrating that the $3$-DOF active spine significantly enhances the robot’s swiftness, maneuverability, precision, and stability.
\end{enumerate}

The remainder of this paper is organized as follows. Section II presents the system design of S-Cheetah. Section III introduces the reinforcement learning method used for control. In Section IV, simulation experiments verify the reliability of the proposed method, followed by a conclusion of the paper in Section V.

\section{System Design}

\subsection{Overview of Structure}

\begin{figure}[t]
    \centering
    \includegraphics[scale=0.27]{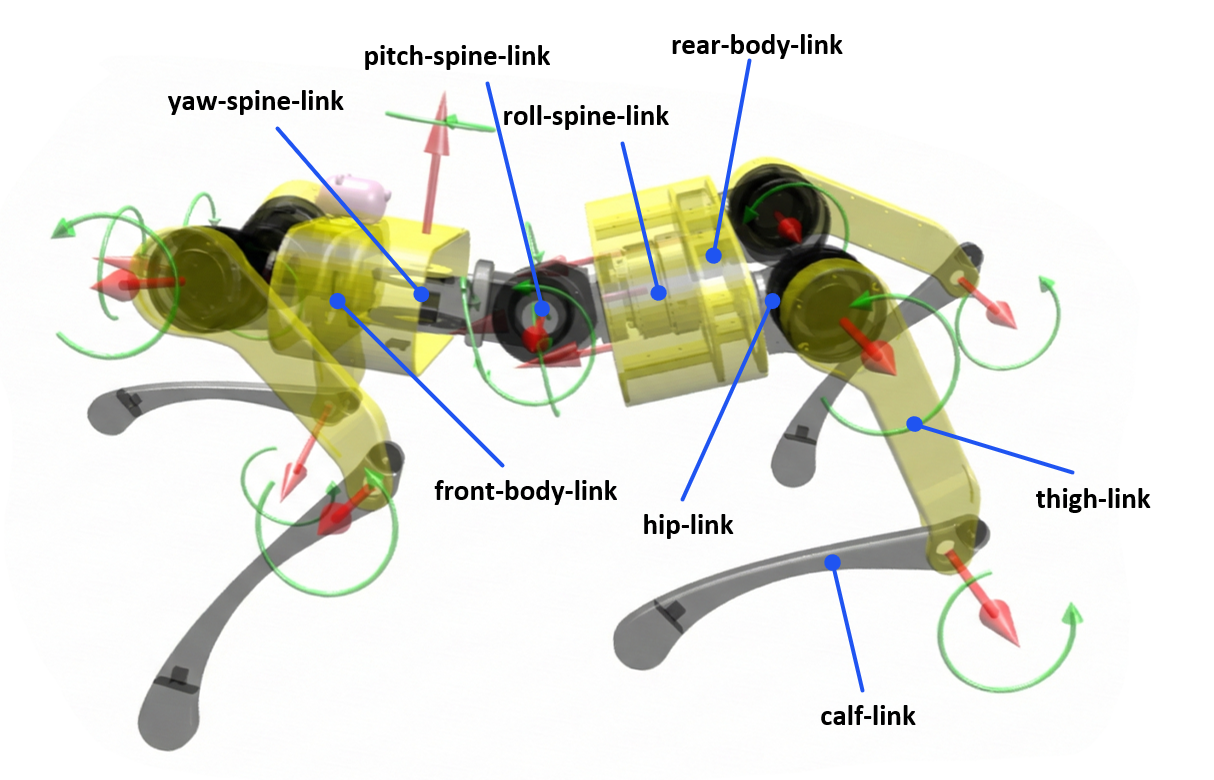}
    \caption{Structural composition and joint rotation representation of S-Cheetah.}
    \label{Voltah_dofs}
\end{figure}

As shown in Fig.~\ref{Voltah_dofs}, the quadrupedal robot S-Cheetah has a total of 15 DOFs, including 3 DOFs for each leg and 3 DOFs for the spine. The rotation axis of each joint is indicated by the red arrows in the figure, while the corresponding rotational motion is illustrated by the green arrows.

The trunk of S-Cheetah consists of two parts, namely the front-body-link and the rear-body-link, which are connected via a spine module. Each leg is composed of a hip-link, a thigh-link, and a calf-link. The robot has a total mass of $20\ \text{kg}$, an overall body length of $625\ \text{mm}$, a width of $380\ \text{mm}$, and a leg length of $580\ \text{mm}$, with a standing height of approximately $440\ \text{mm}$. The peak torque of the leg joint motors is $33.5\ \text{N}\cdot\text{m}$. To more effectively power the robot's locomotion, the spine joint motors are designed with a higher peak torque of $50\ \text{N}\cdot\text{m}$.

\begin{figure}[t]
    \centering
    \includegraphics[scale=0.25]{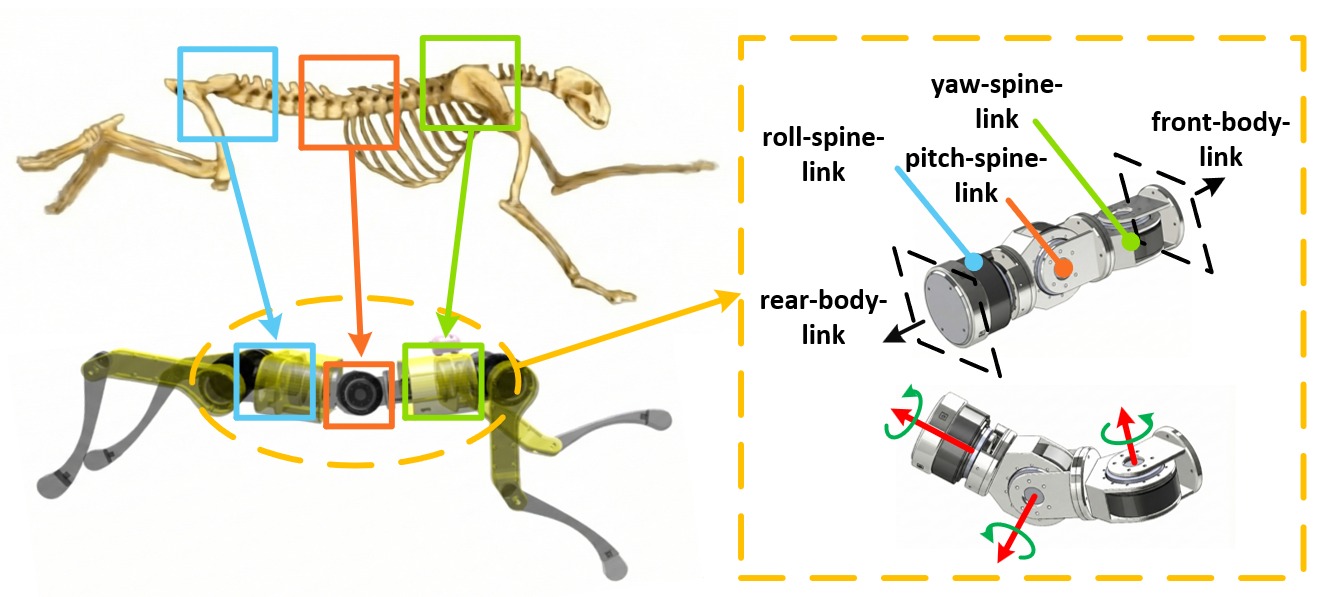}
    \caption{Design of S-Cheetah’s spine joints and their correspondence to a quadrupedal spine.}
    \label{spine}
\end{figure}

\subsection{Design of Spine Joint}
As shown in Fig.~\ref{spine}, the spine module consists of three serially connected joints, arranged from front to rear as the yaw-spine-link, pitch-spine-link, and roll-spine-link. The angular ranges of the yaw, pitch, and roll spine joints are $\pm 1\ \text{rad}$ ($\approx \pm 57.3^{\circ}$), $\pm 1.5\ \text{rad}$ ($\approx \pm 85.9^{\circ}$), and $\pm 2\ \text{rad}$ ($\approx \pm 114.6^{\circ}$), respectively. The ordering of the three serial spine joints is determined based on the following considerations:

\begin{enumerate} 
\item The yaw-spine-link is placed at the foremost position, allowing the front half of the robot to yaw independently. This mimics the gaze-oriented steering behavior observed in animals during prey pursuit, deflecting the forelegs' workspace relative to the hind legs to guide the body naturally into a turn. 
\item The pitch-spine-link is positioned at the middle of the spine module. Placing it at the mechanical center between the fore and hind hip joints maximizes the stride-length gain from spinal flexion and extension, replicating the bow-like elastic energy cycle of the cheetah's lumbar spine. An off-center placement would create unbalanced moment arms, reducing the effectiveness of spinal energy storage. 
\item The roll-spine-link is located at the rear end of the spine module, near the pelvis and hind legs. This allows the posterior trunk to twist relative to the anterior trunk, emulating the independent pelvic adjustments animals perform on uneven terrain and the lateral roll of the hind legs to counteract centrifugal forces during high-speed cornering. 
\end{enumerate}

The introduction of these three serial spine joints enables comprehensive replication of the dexterous spinal motions of quadrupeds in roll, pitch, and yaw. Moreover, the spine configuration characterized by ``guidance prioritized at the front, power centralized in the middle, and compliance at the rear'' allows S-Cheetah to exhibit locomotion dynamics that more closely resemble those of biological quadrupeds.

\section{Reinforcement Learning-Based Control For Agile Locomotion}
\subsection{RL Framework} 

\begin{figure}[t]
    \centering
    \includegraphics[scale=0.275]{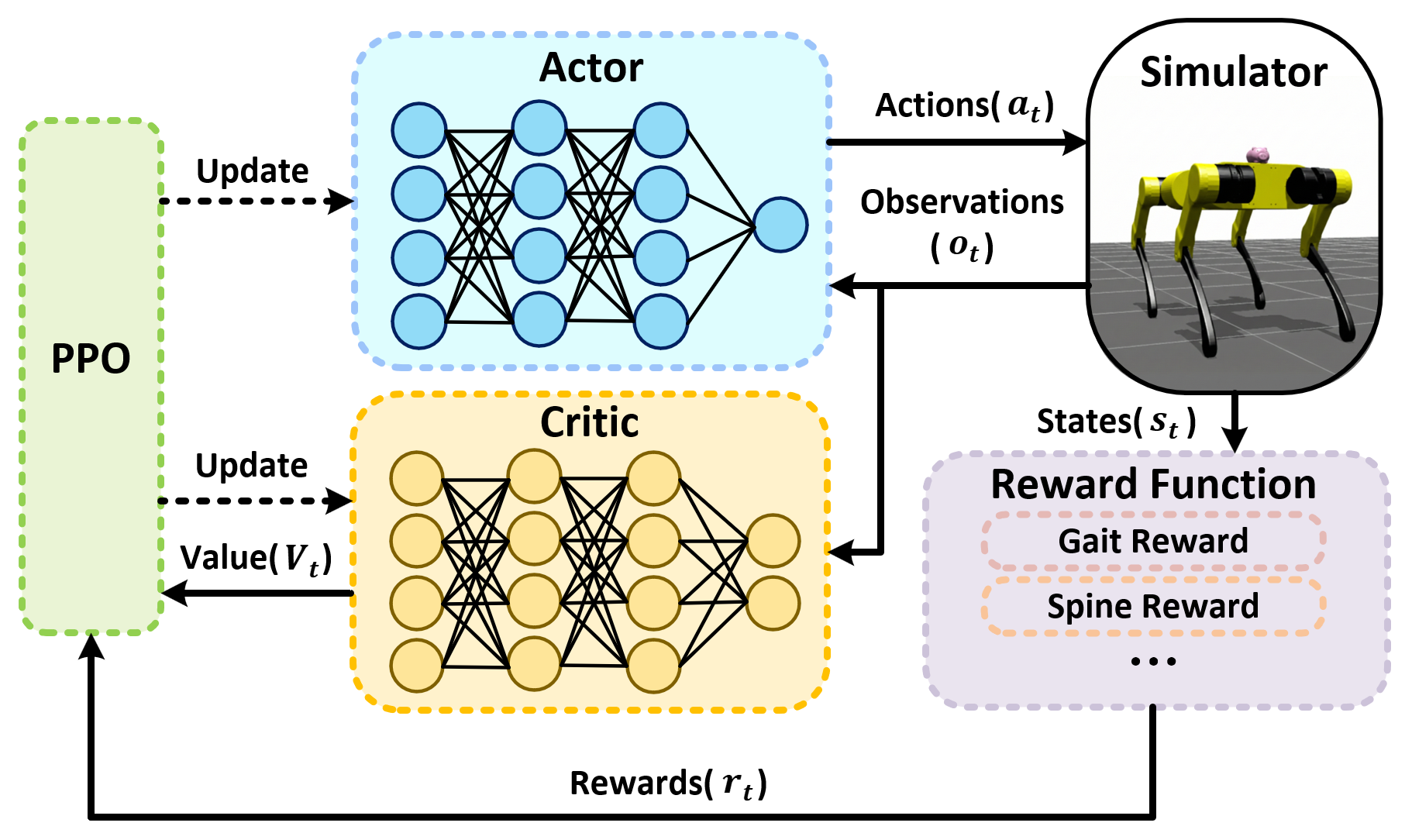}
    \caption{Basic framework of the reinforcement learning method used for training.}
    \label{rl}
\end{figure}

As shown in Fig.~\ref{rl}, the design of the reinforcement learning framework is based on the RSL-RL architecture \cite{schwarke2025rsl}. Two networks, namely the actor and the critic, are trained simultaneously: the actor generates actions, while the critic estimates the value function. Proximal Policy Optimization (PPO) \cite{schulman2017proximal} is adopted as the learning algorithm to update the parameters of both the actor and critic networks.

During control execution, only the actor network is employed. The actor outputs scaled target joint angles for 15 joints (12 leg joints and 3 spine joints) at a frequency of $50\ \text{Hz}$. These target joint angles are then fed into a simple PD position controller, which computes the corresponding joint torques to drive the robot’s motion.

\subsection{Curriculum Learning Design}
Training agile locomotion behaviors such as galloping and turning directly with full-range velocity commands is challenging, as the agent struggles to explore effectively under high-speed targets from the outset. We address this with a two-fold curriculum learning strategy. First, we apply an adaptive velocity range curriculum: the commanded linear and angular velocity ranges are initialized to $10\%$ of their full target ranges and progressively expanded as training performance improves. Specifically, when the mean episodic tracking reward exceeds a predefined threshold ($80\%$ for linear velocity, $60\%$ for angular velocity), the range is incrementally increased by $\Delta=0.1$ per training iteration until the full range is reached. The linear and angular velocity curricula operate independently, allowing the agent to master forward locomotion and yaw control at their own pace.

Second, we introduce an acceleration-constrained command generator to avoid abrupt velocity jumps that the robot cannot physically track. Rather than issuing step-wise target velocities, the commanded velocity $\mathbf{v}_{t}$ is smoothly ramped toward the target $\mathbf{v}^*$ at each simulation step:
\begin{equation}
\mathbf{v}_{t+1}=\mathbf{v}_t+\mathrm{clamp}(\mathbf{v}^*-\mathbf{v}_t,-\mathbf{a}_{\text{max}}\Delta t,\mathbf{a}_{\text{max}}\Delta t),
\end{equation}
where $\Delta t$ is the control time step and $\mathbf{a}_{\text{max}}$ is sampled from a predefined range. For the turning task, the angular velocity command is further delayed by a random interval after the linear velocity command, enabling the robot to first reach a stable forward speed before initiating a turn. This combined curriculum significantly improves training stability and final performance.

\subsection{Main Reward Function Design}

\subsubsection{Gait Reward}
Gallop is the primary gait exhibited by quadrupeds during maximal-speed locomotion. A previous study \cite{11247619} provides a detailed classification of galloping gaits: based on limb-ground contact sequences, gallop is divided into transverse and rotary; based on the number of aerial phases, it is categorized into G0, GG, GE, and G2. Agile quadrupeds like the cheetah typically adopt the rotary G2 gallop, characterized by two aerial phases, as it optimally balances velocity, stability, and energy efficiency. Accordingly, we design a gait reward function to guide the policy toward learning this specific gait for high-speed locomotion.

\begin{figure}[t]
    \centering
    \includegraphics[scale=0.24]{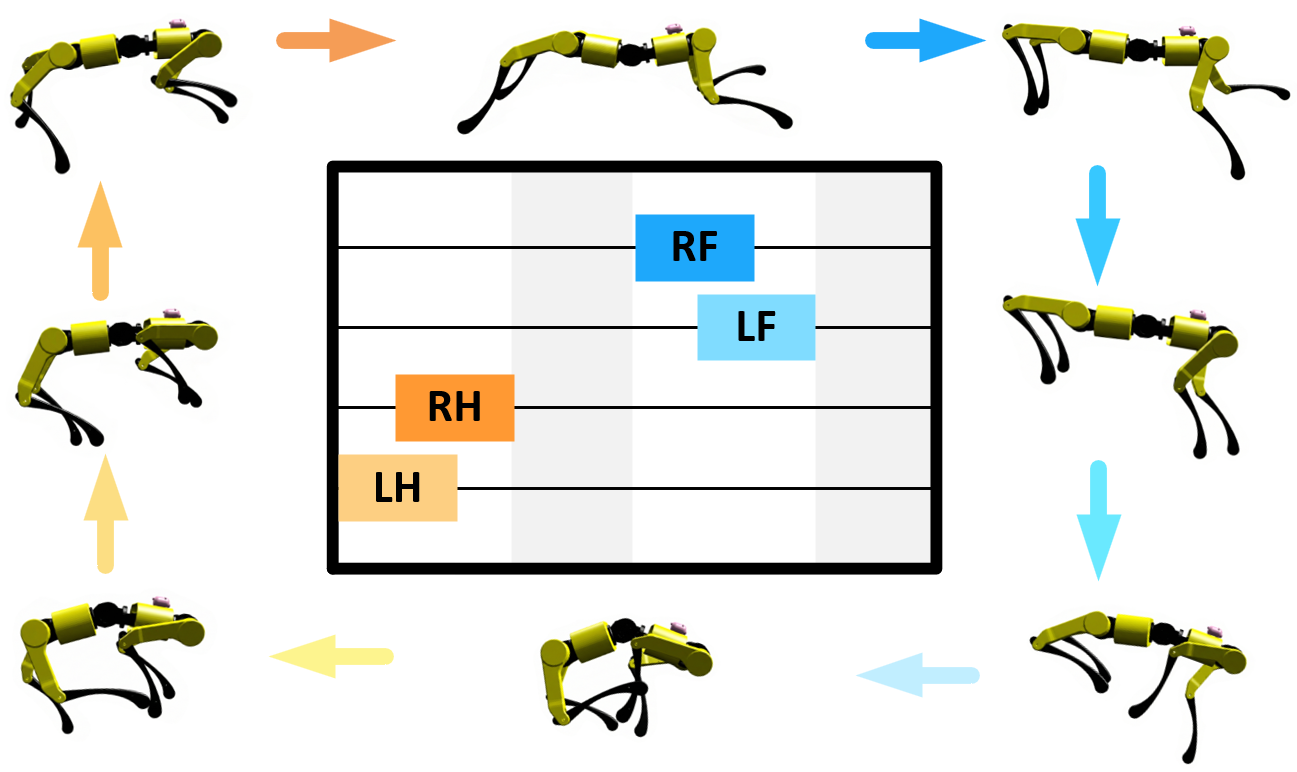}
    \caption{Rotary G2 gallop performance and its gait diagrams.}
    \label{gallop}
    \vspace{-0.3cm}
\end{figure}

Fig.~\ref{gallop} illustrates the locomotion performance and gait diagrams of the rotary G2 gallop. The diagrams depict the temporal sequence of limb–ground contacts (where LH, RH, LF, and RF denote the left/right hind/forelimbs, respectively), with gray segments indicating aerial phases where no limbs touch the ground.

The core idea of our gait reward design is to specify only the minimal footfall pattern distinguishing rotary gallop from other gaits, without prescribing the full gait cycle, duty factor, or stride frequency. In gallop, ipsilateral limb pairs exhibit a characteristic temporal asymmetry: the leading leg contacts the ground slightly before the trailing leg with a non-zero phase offset. This distinguishes gallop from bound, where ipsilateral limbs land simultaneously. To capture this, we define a phase-offset reward based on a Gaussian kernel. For each ipsilateral pair (front and rear), we compute the temporal phase offset between their ground-contact events. To account for the periodic nature of the gait cycle, we calculate the shortest phase distance $\Delta \delta$ between the measured offset and the target offset $\delta^*$. The reward is formulated as:
\begin{equation}
r_{\text{pair}} = \exp\left(-\frac{\Delta \delta^2}{\sigma^2}\right),
\end{equation}
where $\sigma$ controls the tolerance around the target offset.

The total gait reward is the product of the front and rear pair rewards: $r_{\text{gait}}=r_{\text{front}}\cdot r_{\text{rear}}$. Crucially, we impose no explicit constraint on the phase relationship between the front and rear limb groups, nor on the duration of aerial phases. By setting different target offsets for the front ($\delta_{f}^*$) and rear ($\delta_{r}^*$) pairs, the reward provides sufficient guidance to prevent bounding ($\delta^*=0$) while preserving flexibility. This deliberate under-specification, combined with auxiliary rewards like air time and velocity tracking, allows the policy to naturally discover the full rotary G2 gallop---including its two distinct aerial phases and characteristic circular footfall sequence (LH→RH→RF→LF)---as an emergent optimal solution.

\subsubsection{Spine Undulation Reward}
In prior work \cite{8956332}, no dedicated reward was designed to actively drive spinal joint motion; instead, the spine was expected to learn useful movements implicitly through velocity tracking rewards and power penalties. While this passive approach does produce some spinal oscillation during locomotion, it is highly susceptible to reward hacking; the policy tends to converge to minimal or irregular spinal movements that satisfy the tracking objective without leveraging the spine to enhance propulsion. In nature, agile quadrupeds utilize a rhythmic, wave-like spinal motion known as undulation to maximize stride length and generate explosive forward thrust. To induce this crucial biomechanical feature, we design an explicit spine undulation reward for the pitch spinal joint. This reward encourages large-amplitude, phase-coordinated spinal flexion-extension to actively power high-speed galloping.

The reward consists of three components. First, a phase coordination term evaluates whether the spine moves in the biomechanically correct direction relative to the limbs. We compute an effective leg velocity $\bar{v}_{\text{leg}}$ by averaging the rear thigh velocities with the sign-inverted front thigh velocities, so that a positive value consistently indicates a phase requiring spinal extension. The phase score is then:
\begin{equation}
r_{\text{phase}} = \tanh(\dot{\theta}_{\text{pitch}} \cdot \bar{v}_{\text{leg}} \cdot \alpha),
\end{equation}
where $\dot{\theta}_{\text{pitch}}$ is the spine angular velocity and $\alpha$ is a sensitivity coefficient. 

Second, an amplitude incentive encourages the spine to reach a target deflection angle. Based on the sign of $\bar{v}_{\text{leg}}$, the expected spine state $s\in\{-1,+1\}$ is determined (flexion or extension), and the amplitude reward is computed as:
\begin{equation}
r_{\text{amp}} = s \cdot \theta_{\text{pitch}}.
\end{equation}

During quadrupedal locomotion, there is a substantial disparity in the spinal angular limits between flexion and extension \cite{belyaev2024running}. Typically, the maximum angle achieved during flexion significantly exceeds that of extension. Consequently, we define an expected threshold $\theta_{\text{th}}$ (which denotes either $\theta_{\text{flex}}$ or $\theta_{\text{ext}}$ depending on the spine state) to faithfully simulate the spinal dynamics. The effective amplitude reward is then clamped as $\bar{r}_{\text{amp}} = \mathrm{clamp}(r_{\text{amp}},\ 0,\ \theta_{\text{th}})$.

Third, an excess penalty prevents over-rotation beyond the desired range:
\begin{equation}
p=w_{e}(\max(0,\ r_{\text{amp}}-\theta_{\text{th}}))^{2},
\end{equation}
where $w_{e}$ is the excess penalty.

The final reward combines these as:

\begin{equation}
R_{\text{spine\_undulation}} = r_{\text{phase}} \cdot (1 + w_{b} \cdot \bar{r}_{\text{amp}}) - p,
\end{equation}
where $w_{b}$ is the amplitude boost weight and $\bar{r}_{\text{amp}}$ is the clamped amplitude reward. This design ensures that the spine is rewarded for moving in the correct phase with sufficient amplitude, while being penalized for exceeding safe mechanical limits.

\subsubsection{Spine Steering Reward}
Conventional rigid-trunk quadrupedal robots can only achieve turning through differential leg velocities between the left and right sides, making it inherently difficult to decouple forward locomotion from yaw maneuvers. The introduction of a yaw-axis spinal joint addresses this limitation, functioning analogously to the steering mechanism of a vehicle that directs the front axle. To encourage the policy to actively utilize this joint for turning, we design a spine steering alignment reward.

The reward incentivizes the yaw spine joint to deflect in the direction consistent with the commanded angular velocity. To ensure that the spine is only activated during intentional turning maneuvers, we incorporate a dead-zone threshold $\omega_{\text{th}}$ into a piecewise reward function:
\begin{equation}
R_{\text{spine\_steering}} = 
\begin{cases} 
0, & \text{if } |\omega_z^*| < \omega_{\text{th}} \\
\tanh\left(\frac{-\theta_{\text{yaw}}}{\omega_z^*} \cdot k \right), & \text{otherwise}
\end{cases},
\end{equation}
where $k$ is a scaling factor controlling how quickly the reward saturates. When the yaw rate command is within the dead-zone ($|\omega_z^*| < \omega_{\text{th}}$), the reward is strictly zeroed out, effectively preventing any spurious spine deflection during straight-line locomotion. When an active turn is commanded ($|\omega_z^*| \ge \omega_{\text{th}}$), the reward evaluates the alignment. Crucially, the formulation inherently encapsulates the sign of the command in the denominator $\omega_z^*$. This ensures that a positive reward is given when the spine deflects in the correct steering direction, while an incorrect deflection yields a negative penalty, providing a clear directional learning signal.

\section{Experiments}
To validate the effectiveness of the proposed spinal quadrupedal robot and reward design, we conducted extensive experiments across two physics engines. Our training framework, built upon the open-source codebase \cite{fan-ziqi2024robot_lab}, leverages NVIDIA Isaac Sim to enable massively parallel, GPU-accelerated reinforcement learning. The learned policy was first evaluated in Isaac Sim and then directly transferred to MuJoCo---a simulator favored in the legged robotics community for its high-fidelity contact dynamics and minimal sim-to-real gap. This seamless cross-engine transfer, achieved without fine-tuning or domain adaptation, underscores the policy’s robustness and provides encouraging evidence for future physical deployment.

To isolate the contribution of the articulated spine, we compared two configurations: (1) the full robot model with a 3-DOF active spine, and (2) a rigid-trunk baseline with locked spinal joints. Both configurations shared identical hyperparameters, observation/action spaces, and reward functions (excluding spine-specific terms). Furthermore, both were trained for a consistent number of iterations to ensure a fair and rigorous comparison.

\subsection{Spine-Empowered High-Speed Galloping}
The rotary G2 gallop, the fastest gait employed by agile quadrupeds like cheetahs, relies on coordinated spinal flexion and extension to augment limb propulsion. As illustrated in Fig.~\ref{voltah}(a), our proposed framework successfully realizes this complex asymmetric gait, effectively leveraging spinal dynamics to enhance locomotor performance.

\begin{figure}[t]
    \centering
    \includegraphics[scale=0.14]{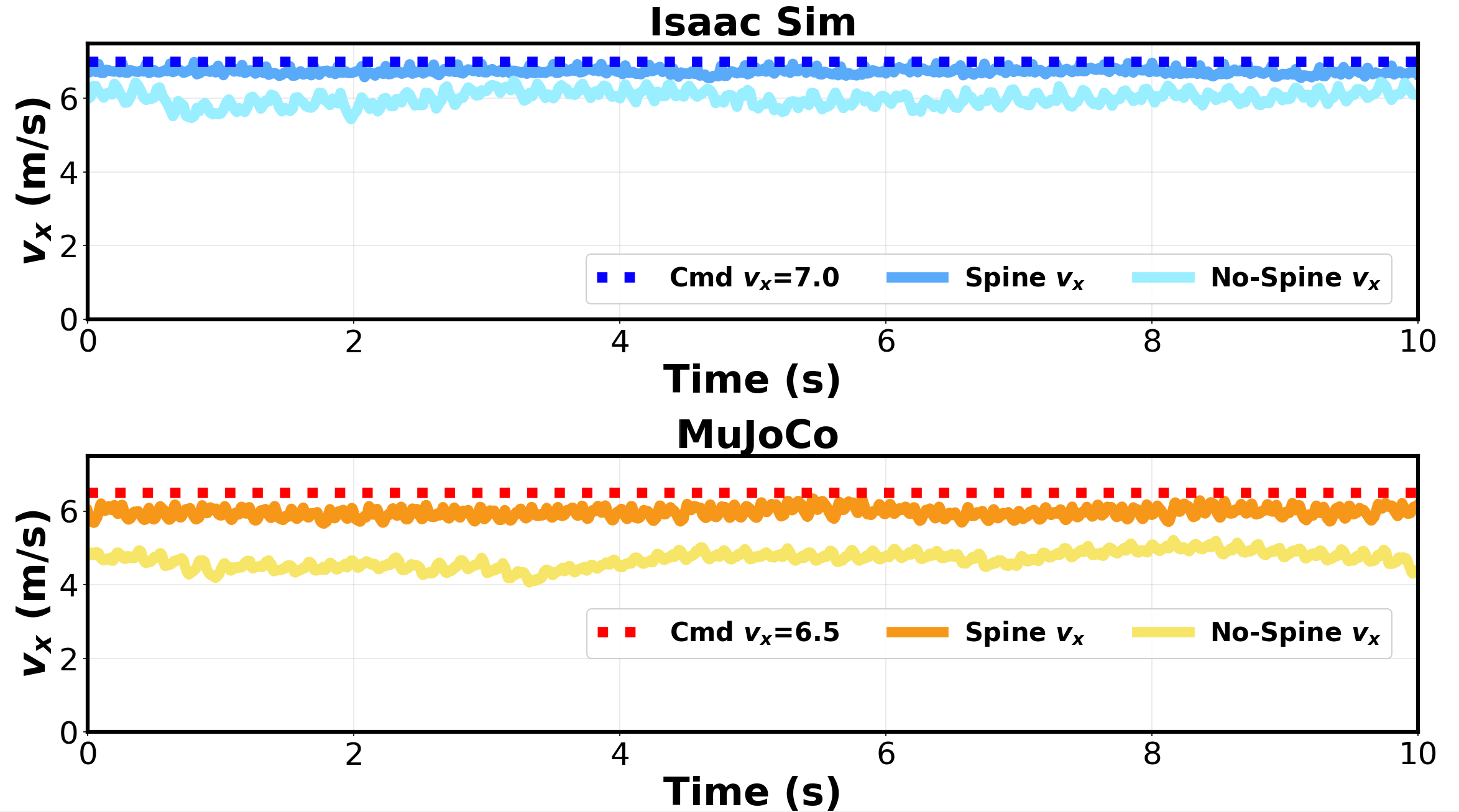}
    \caption{The $v_{x}$ tracking performance during galloping.}
    \label{data_gallop}
    \vspace{-0.2cm}
\end{figure}

To evaluate speed limits, we applied high-velocity tracking commands across both simulators (Fig.~\ref{data_gallop}). In Isaac Sim, the active-spine robot achieved a stable speed of $6.9\ \text{m/s}$, representing a $15\%$ improvement over the $6.0\ \text{m/s}$ rigid-trunk baseline. In MuJoCo, the performance gain was even more pronounced: the active-spine robot reached $6.0\ \text{m/s}$, representing a $25\%$ enhancement over the baseline's $4.8\ \text{m/s}$. Notably, both results significantly exceed the current state-of-the-art (SOTA) for simulated spined quadrupeds, which stands at $4.9\ \text{m/s}$ \cite{10907590}.

\begin{figure}[h]
    \centering
    \includegraphics[scale=0.215]{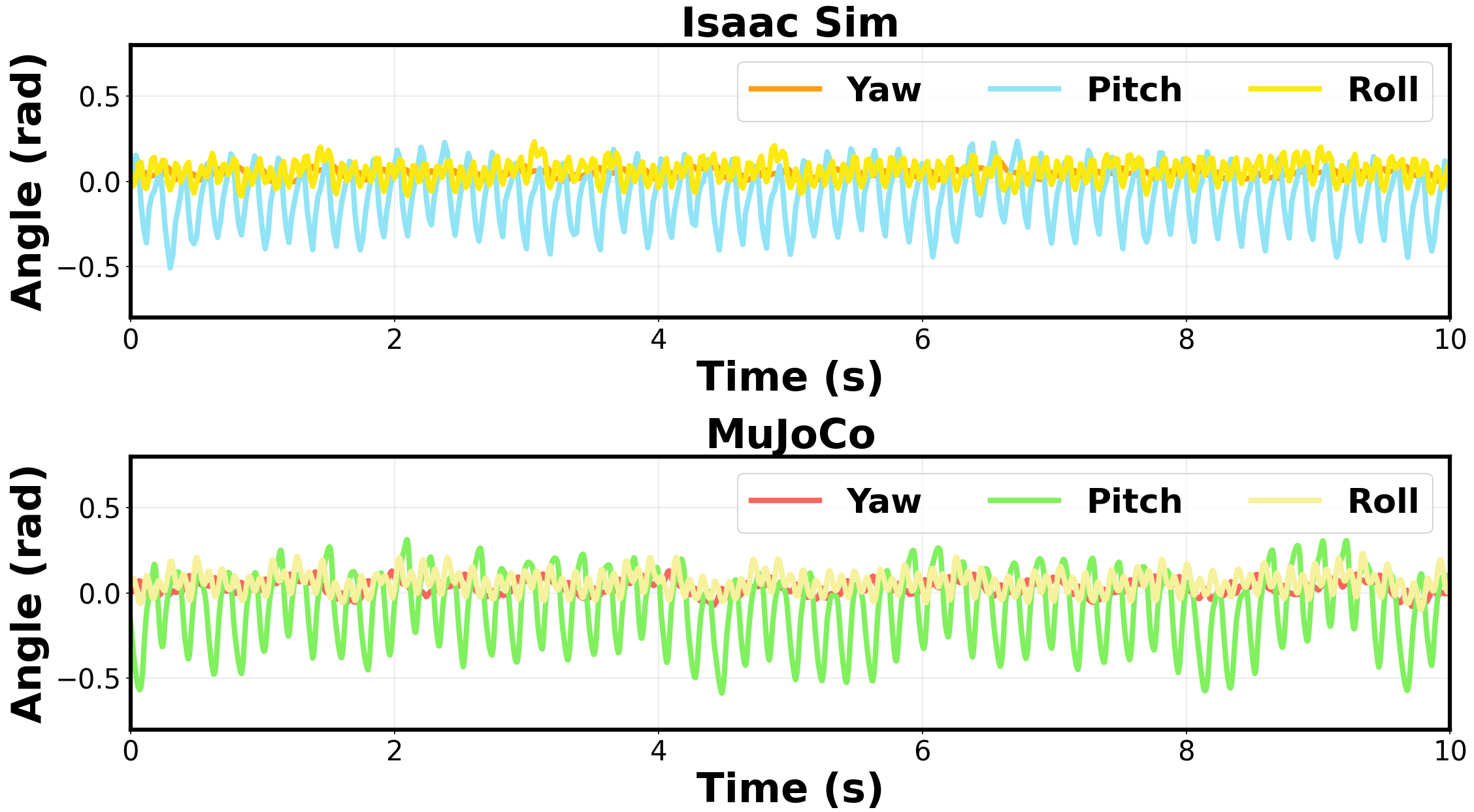}
    \caption{Angular variations of the three spine joints during galloping.}
    \label{data_gallop_pos}
    \vspace{-0.5cm}
\end{figure}

Kinematic analysis of the spinal joints in Fig.~\ref{data_gallop_pos} reveals that the pitch joint exhibits substantial asymmetric variations, with angular ranges of $[-0.5,\ 0.25]\ \text{rad}$ in Isaac Sim and $[-0.6,\ 0.3]\ \text{rad}$ in MuJoCo. This behavior aligns with biological principles where spinal flexion magnitude is approximately double that of extension during galloping. Meanwhile, the roll joint maintains minimal variation, ensuring lateral stability. This bio-inspired asymmetry contrasts sharply with existing MPC-based methods \cite{10160896}, which prescribe symmetric spinal motion (e.g., $[-0.42,\ 0.42]\ \text{rad}$), a simplification that contradicts the natural kinematic characteristics of rapid galloping.

\subsection{Spine-Assisted Agile Turning}
\subsubsection{Turning Capability Evaluation}
To evaluate the turning enhancements enabled by the articulated spine, we conducted angular velocity tracking experiments in both simulators. We measured the maximum yaw rate ($\omega_z$) for the active-spine and rigid-trunk configurations under two conditions: in-place turning and moving turning (tracking a constant forward velocity of $1\ \text{m/s}$), as depicted in Fig.~\ref{voltah}(b).

\begin{figure}[t]
    \centering
    \includegraphics[scale=0.19]{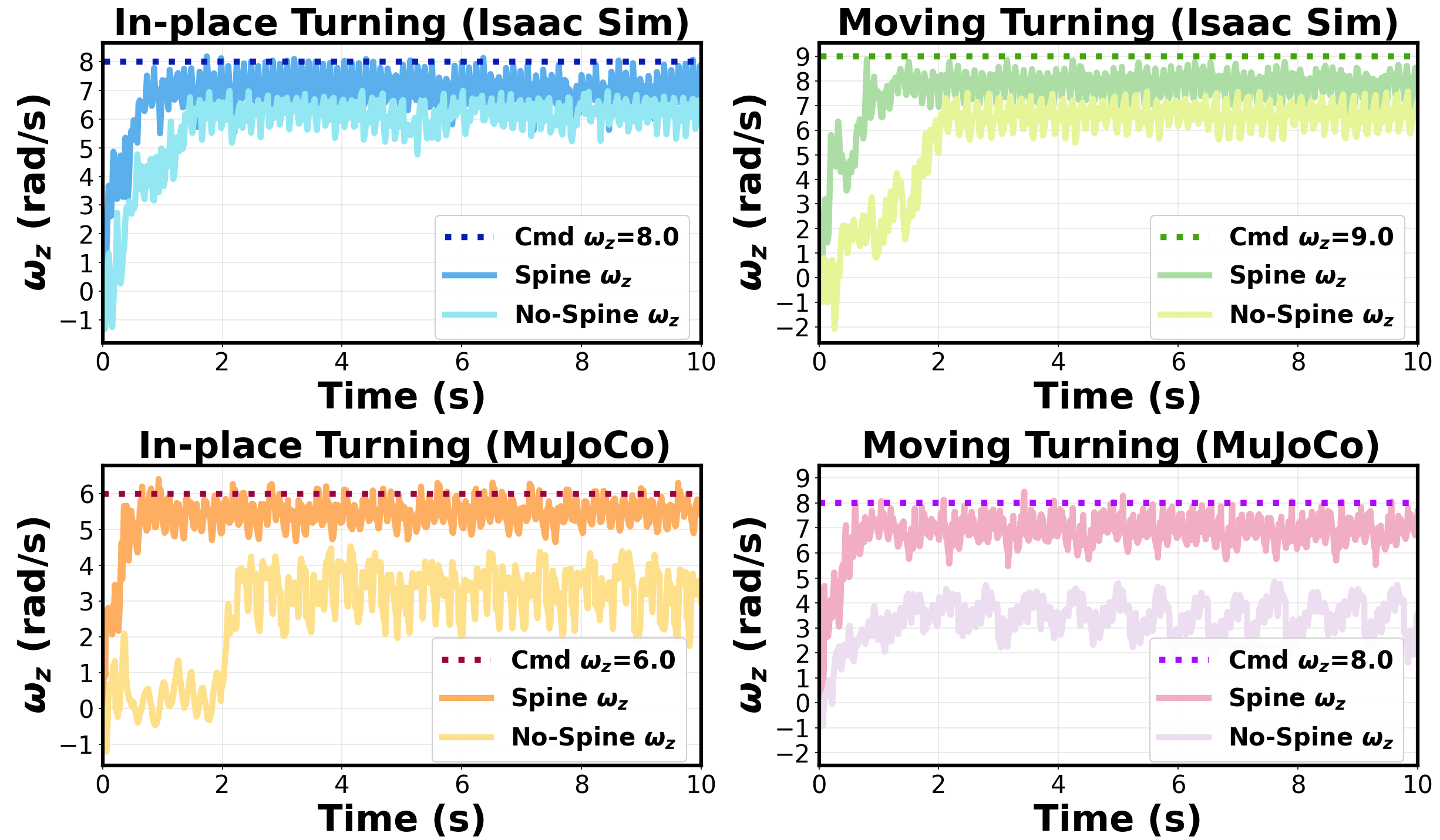}
    \caption{The $\omega_z$ tracking performance during turning.}
    \label{data_turn}
    \vspace{-0.2cm}
\end{figure}

\begin{figure}[h]
    \centering
    \includegraphics[scale=0.2]{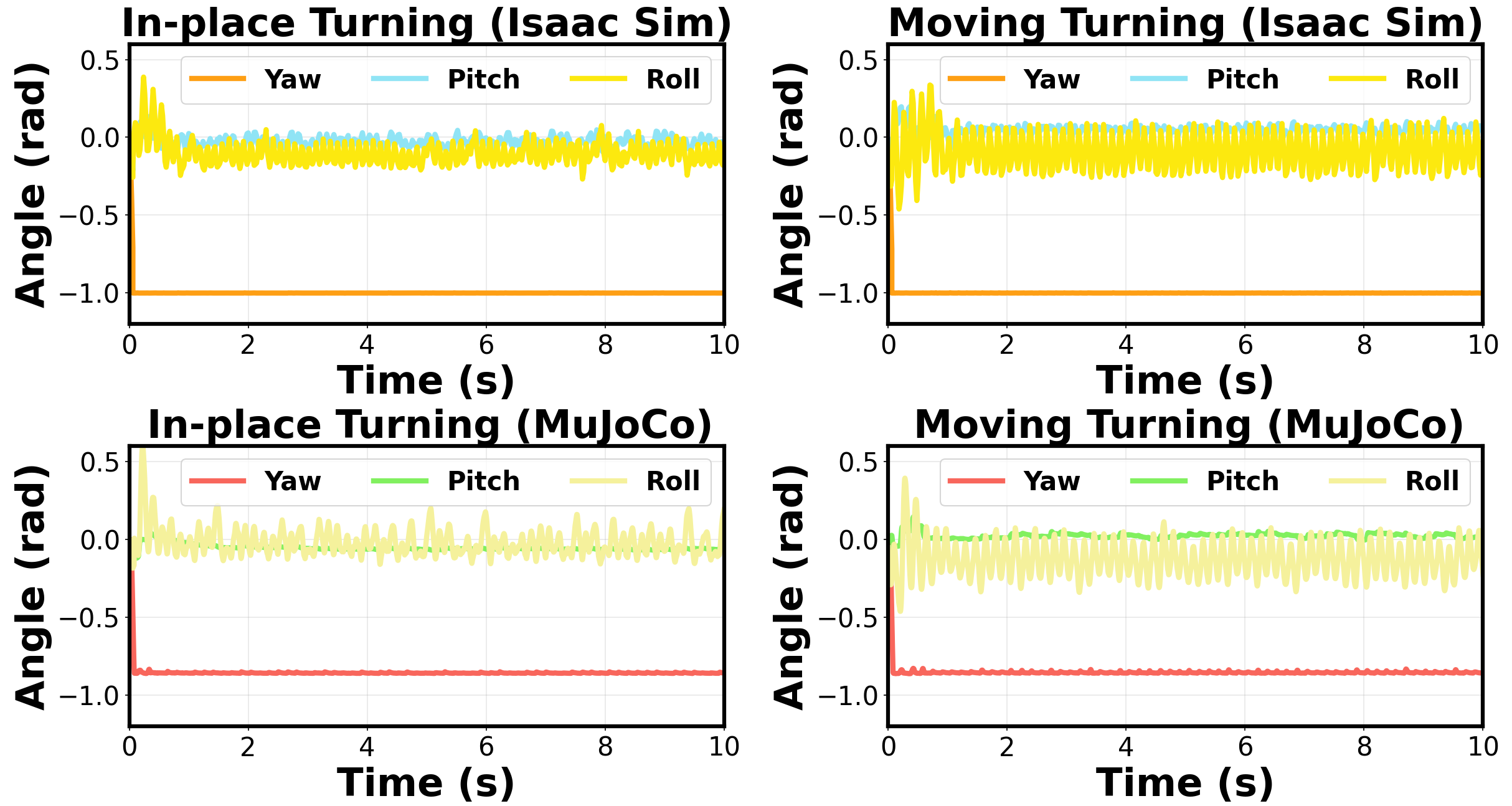}
    \caption{Angular variations of the three spine joints during turning.}
    \label{data_turn_pos}
    \vspace{-0.5cm}
\end{figure}

\begin{figure*}[h]
    \centering
    \includegraphics[scale=0.57]{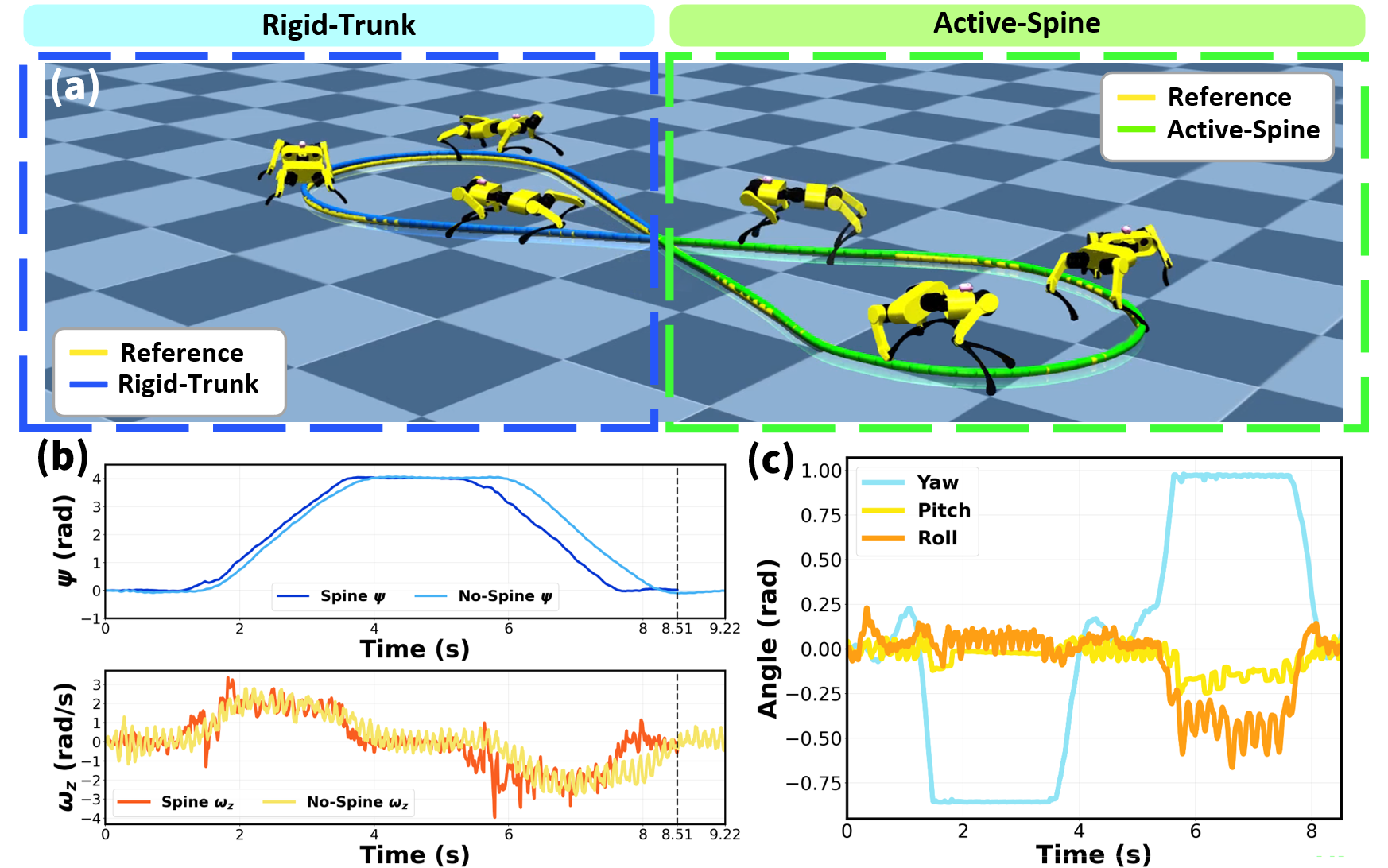}
    \vspace{-0.2cm}
    \caption{Evaluation of the ``figure-8'' path-tracking performance. (a) Spatial trajectories of the active-spine and rigid-trunk robots. (b) Variations in yaw angle $\psi$ and yaw rate $\omega_z$ for both configurations. (c) Angular variations of the three spinal joints during path-tracking.}
    \label{data_turn_8}
    \vspace{-0.5cm}
\end{figure*}

As shown in Fig.~\ref{data_turn}, the active-spine configuration consistently and significantly outperformed the rigid-trunk baseline. During in-place turning, the active-spine robot achieved a stable maximum $\omega_z$ of $7.2~\mathrm{rad/s}$ in Isaac Sim, representing a $16.1\%$ improvement over the rigid baseline, and $5.3~\mathrm{rad/s}$ in MuJoCo, a $60.1\%$ increase compared to its rigid counterpart. The performance gap further widened during moving turning tasks: the active-spine robot reached $7.8~\mathrm{rad/s}$ in Isaac Sim, marking a $22.9\%$ gain over the rigid baseline, and $6.6~\mathrm{rad/s}$ in MuJoCo, an impressive $83.3\%$ enhancement relative to the rigid-trunk model. Notably, these yaw rates substantially surpass the $3~\mathrm{rad/s}$ state-of-the-art previously reported for quadrupeds with yaw-actuated spines \cite{10160896}.

Kinematic analysis of the spinal joints in Fig.~\ref{data_turn_pos} reveals the underlying mechanisms driving this agility. During high-speed turns, the spinal yaw joint operates near its mechanical limits to maximize rotational capability. Concurrently, the spinal roll joint actively adjusts to execute a banking maneuver---leaning into the curve. This bio-inspired compensation effectively counteracts centrifugal forces, significantly enhancing the robot's dynamic stability during rapid directional changes.

\subsubsection{Path Tracking Performance}
To further investigate how the 3-DOF spine enhances maneuverability, we conducted a ``figure-8'' path-tracking task in MuJoCo. The trajectory comprises two circular arcs ($R=1\text{ m}$) connected by intersecting straight segments ($L=4\text{ m}$). During the task, the robot maintains a constant commanded forward velocity $v_x$, while a proportional (P) controller regulates steering by translating heading errors into target yaw rates $\omega_z$ for the policy to track.

As illustrated by the spatial trajectories in Fig.~\ref{data_turn_8}(a), the active-spine robot tracks the reference path more precisely than the rigid-trunk baseline, leveraging its mid-body DOF for fluid steering. Taking the $v_x = 2.0\text{ m/s}$ case as a representative example, quantitative analysis in Fig.~\ref{data_turn_8}(b) demonstrates that the active-spine configuration generates a higher transient yaw rate when entering curves, enabling more rapid heading adjustments than the rigid-trunk baseline. The spinal kinematic profiles in Fig.~\ref{data_turn_8}(c) further reveal this coordination: as the robot transitions into turns, the spinal yaw angle increases to tighten the turning radius, while the roll angle simultaneously adjusts to provide banking support.

Table~\ref{tab:performance_table_merged} summarizes the performance metrics across various velocities. The active-spine robot consistently outperforms the rigid-trunk counterpart, achieving shorter completion times while maintaining superior accuracy in both path following and velocity tracking. Notably, these consistent improvements are achieved under a straightforward tracking framework with identical velocity commands, rather than using a specialized planner trained to push the physical limits of either platform. While beyond the scope of this work, we anticipate that a dedicated  time-optimal policy could further magnify this performance gap, as the  active spine provides greater mechanical headroom for extreme maneuvers.

\begin{table}[htbp]
    \vspace{-0.3cm}
    \fontsize{6pt}{8pt}\selectfont
    \setlength{\tabcolsep}{3pt}
    \centering
    \caption{Path-tracking performance of rigid-trunk and active-spine robots at various velocities $v_{x}$.}
    \label{tab:performance_table_merged}
    \begin{tabular}{|c|cc|cc|cc|}
    \hline
    \multirow{2}{*}{\makecell{$v_x$ \\ (m/s)}}
    & \multicolumn{2}{c|}{Total Time (s)}
    & \multicolumn{2}{c|}{\makecell{Average Path \\ Tracking Error (m)}}
    & \multicolumn{2}{c|}{\makecell{Average Velocity \\ Tracking Error (m/s)}} \\
    \cline{2-7}
    & {Rigid-Trunk} & {Active-Spine}
    & {Rigid-Trunk} & {Active-Spine}
    & {Rigid-Trunk} & {Active-Spine} \\
    \hline
    1.0 & 16.995 & 14.695 & 0.039 & 0.036 & 0.118 & 0.118 \\
    1.5 & 11.655 & 10.535 & 0.037 & 0.035 & 0.187 & 0.113 \\
    2.0 &  9.220 &  8.510 & 0.044 & 0.034 & 0.251 & 0.139 \\
    2.5 &  8.000 &  7.565 & 0.059 & 0.046 & 0.457 & 0.393 \\
    \hline
    \end{tabular}
    \vspace{-0.5cm}
\end{table}


\subsection{Spine-Driven Aerial Self-Righting}
During the training process, we observed an emergent capability in the active-spine robot: a markedly higher success rate in achieving stable four-point 
landings from random initial poses, a task where the rigid-trunk counterpart frequently suffered body-ground collisions. To systematically evaluate this, we designed an aerial drop experiment from a height of $2\text{ m}$ under three extreme initial orientations: $(\text{pitch}=90^\circ)$, $(\text{roll}=135^\circ, \text{pitch}=45^\circ)$, and a complete inversion $(\text{roll}=180^\circ)$.

\begin{figure}[t]
    \centering
    \includegraphics[scale=0.19]{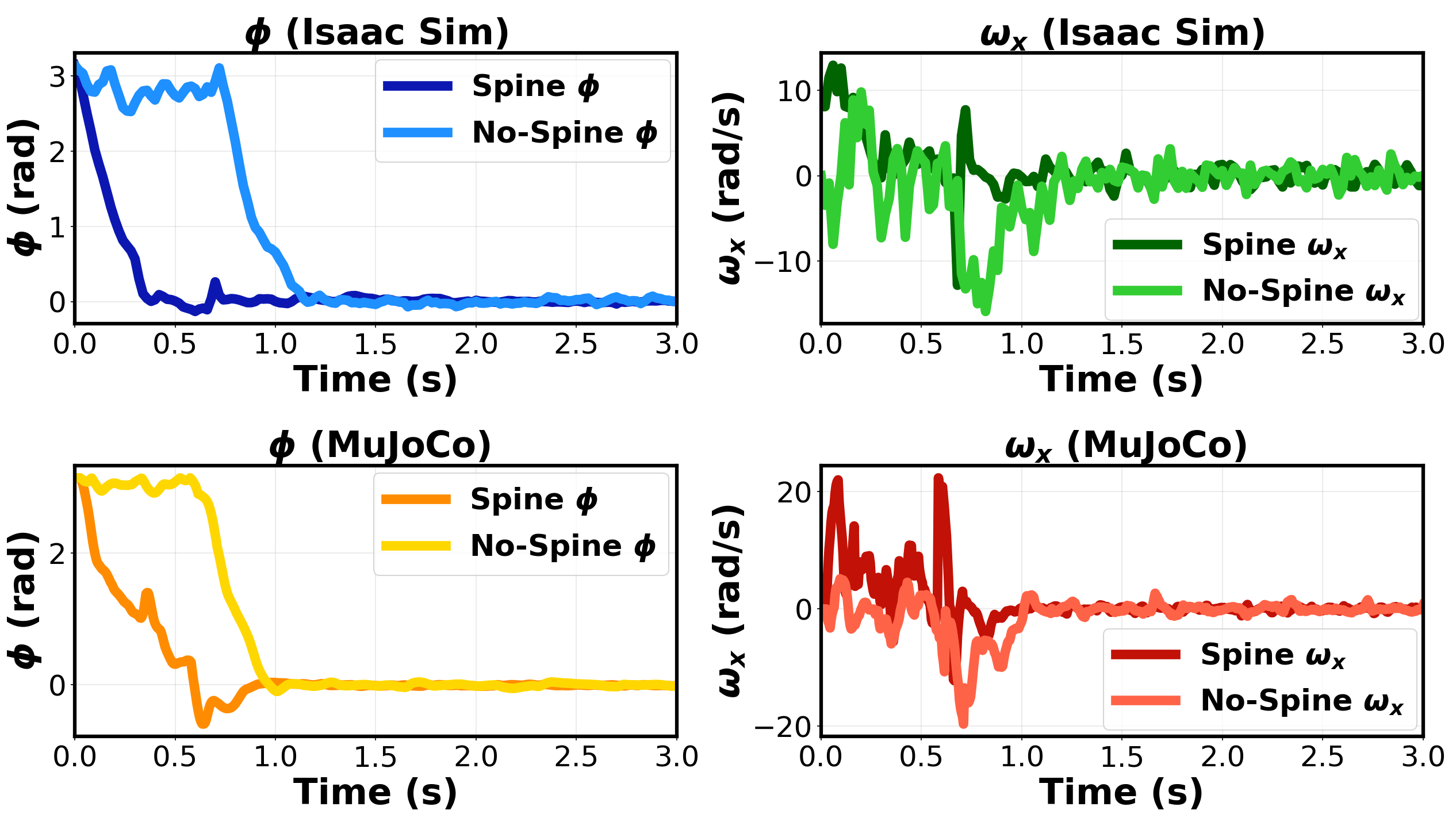}
    \caption{Variations in roll angle and roll rate during aerial self-righting.}
    \label{twist_comp}
    \vspace{-0.3cm}
\end{figure}

\begin{figure}[t]
    \centering
    \includegraphics[scale=0.21]{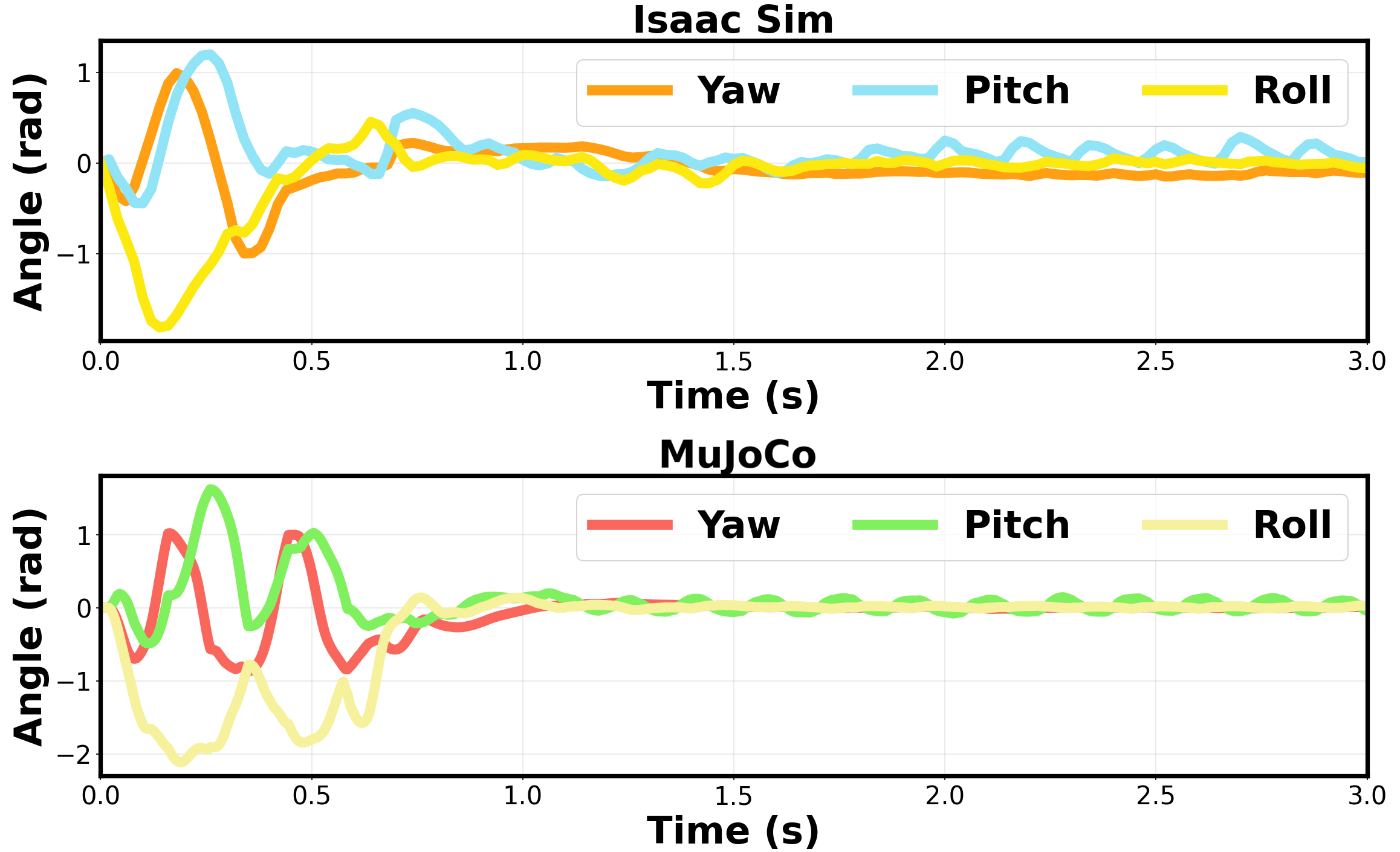}
    \caption{Angular variations of the three spine joints during the aerial self-righting.}
    \label{data_twist}
    \vspace{-0.5cm}
\end{figure}

As illustrated in Fig.~\ref{voltah}(c), the active-spine robot leverages its 3-DOF spinal joints to rapidly modulate its body orientation during descent. Taking the complete inversion ($\phi=180^\circ$) as a representative case, the robot coordinates its spinal rotations to reorient its chassis mid-air, ensuring its feet are optimally positioned for impact, closely mimicking the feline self-righting reflex. Quantitative analysis in Fig.~\ref{twist_comp} confirms that while the rigid-trunk robot fails to significantly modulate its roll angle $\phi$ due to a lack of internal momentum manipulation, the active-spine robot generates substantial roll rate $\omega_x$ to correct its posture, driven by the coordinated joint-angle trajectories shown in Fig.~\ref{data_twist}. Conversely, the rigid-trunk model often impacts the ground on its back, relying solely on post-landing limb recovery.

Table~\ref{tab:success_rates} summarizes the landing success rates across both simulators. Under the most challenging inverted orientation ($\phi=180^\circ$), the rigid-trunk robot exhibits a near-zero success rate. In stark contrast, the active-spine robot achieves near-$100\%$ success in Isaac Sim and $88\%$ in MuJoCo, demonstrating that the articulated spine is indispensable for robust aerial autonomy and landing stability.

\begin{table}[htbp]
    \vspace{-0.3cm}
    \footnotesize
    \centering
    \caption{Success rates (\%) under different initial orientations in two simulation environments.}
    \label{tab:success_rates}
    \setlength{\tabcolsep}{1pt}
    \begin{tabular}{c c c c c}
    \toprule
    \diagbox{Init.\ Orient}{Robot Cfg}
    & \makecell{Rigid-trunk\\(Isaac Sim)}
    & \makecell{Active-spine\\(Isaac Sim)}
    & \makecell{Rigid-trunk\\(MuJoCo)}
    & \makecell{Active-spine\\(MuJoCo)} \\
    \midrule
    $p=90^\circ$ & $60$ & $\bm{>98}$ & $5$  & $\bm{96}$ \\
    $r=135^\circ,\ p=45^\circ$ & $25$ & $\bm{>98}$ & $28$ & $\bm{90}$ \\
    $r=180^\circ$ & $<1$ & $\bm{>98}$ & $<1$ & $\bm{88}$ \\
    \bottomrule
    \end{tabular}
    \vspace{-0.5cm}
\end{table}


\section{Conclusion}
In this study, we introduce S-Cheetah, a quadrupedal robot featuring a bio-inspired serial 3-DOF active spine, which can effectively  replicate the tri-axial rotation of biological quadrupeds. By leveraging a specialized reinforcement learning framework with tailored reward functions, the robot achieves agile locomotion, including high-speed galloping, maneuverable turning, precise path-tracking, and emergent feline-inspired aerial self-righting. Through comprehensive cross-validation in Isaac Sim and MuJoCo, we established a highly robust policy with a minimal sim-to-sim gap. In these simulated environments, S-Cheetah achieved a peak running speed of $6.9\text{ m/s}$ using the rotary G2 gallop gait and an in-place turning rate of $7.2\text{ rad/s}$, substantially outperforming rigid-trunk baselines and current state-of-the-art simulated spinal quadrupedal robots. Crucially, our analysis confirms that the integrated 3-DOF spine significantly enhances the robot’s swiftness, maneuverability, precision, and stability. 

Although this study currently focuses on the design and simulation phases prior to hardware deployment, it provides strong computational evidence that a bio-inspired serial 3-DOF active spine can empower quadrupedal robots with a level of locomotion that closely rivals the agility of natural quadrupeds. Future work will focus on the hardware realization of S-Cheetah for sim-to-real transfer, followed by the integration of imitation learning with canine motion capture (mocap) data to further investigate natural and efficient spinal motion patterns.

\bibliographystyle{./IEEEtran}
\bibliography{references.bib}

\end{document}